# A Wireless Collaborated Inference Acceleration Framework for Plant Disease Recognition

Hele Zhu[1,#], Xinyi Huang[2,#], Haojia Gao[3], Mengfei Jiang[2], Haohua Que[4] and Lei Mu[1,✉]

[1] Southwest Minzu University, Chengdu, Sichuan, China
[2] Beihang University, Beijing, Beijing, China
[3] Beijing University of Technology, Beijing, Beijing, China
[4] Beijing Forestry University, Beijing, Beijing, China
truemoller@outlook.com

**Abstract.** Plant disease is a critical factor affecting agricultural production. Traditional manual recognition methods face significant drawbacks, including low accuracy, high costs, and inefficiency. Deep learning techniques have demonstrated significant benefits in identifying plant diseases, but they still face challenges such as inference delays and high energy consumption. Deep learning algorithms are difficult to run on resource-limited embedded devices. Offloading these models to cloud servers is confronted with the restriction of communication bandwidth, and all of these factors will influence the inference's efficiency. We propose a collaborative inference framework for recognizing plant diseases between edge devices and cloud servers to enhance inference speed. The DNN model for plant disease recognition is pruned through deep reinforcement learning to improve the inference speed and reduce energy consumption. Then the optimal split point is determined by a greedy strategy to achieve the best collaborated inference acceleration. Finally, the system for collaborative inference acceleration in plant disease recognition has been implemented using Gradio to facilitate friendly human-machine interaction. Experiments indicate that the proposed collaborative inference framework significantly increases inference speed while maintaining acceptable recognition accuracy, offering a novel solution for rapidly diagnosing and preventing plant diseases.

## 1 Introduction

Plant diseases can significantly hinder plant growth and physiological functions, leading to a reduction in the quality and yield of agricultural products, resulting in substantial economic losses for farmers. The Deep Neural Network (DNN) [1] has shown tremendous superiority in plant disease identification compared to traditional manual disease identification techniques and is gradually becoming a focal point of research. The DNN-based plant disease recognition methods are mainly categorized into two types: edge-side inference and server-side inference. The former suffers from limited computational power on edge devices, leading to slow inference speeds and high energy consumption, while the latter is constrained by network bandwidth and latency, resulting in significant communication overhead.

To address the issues mentioned above more effectively, this paper presents a framework for collaborative inference acceleration of plant disease recognition. The contributions are as follows: 1) A deep reinforcement learning framework optimizes layer-wise sparsity and network partitioning via greedy algorithms to accelerate collaborative inference. 2) Experiments confirm its effectiveness in enhancing plant disease recognition accuracy. 3) Deployed via a Gradio-based system, it automates disease diagnosis and treatment suggestions through uploaded photos or videos.

## 2 Related Works

Model compression can reduce the size of neural network models, thereby decreasing computation time and achieving inference acceleration. Abdul [2] designed a 28KB int8-quantized model for IoT deployment, enabling real-time diagnosis of 9 plant diseases. Song [3] enhanced CSPDarkNet53[4] with a deep separable convolution, balancing high accuracy and speed in rice disease detection. He [5] introduced deep reinforcement learning for layer-wise pruning, achieving 2.7% higher accuracy, 4× fewer FLOPs, and 1.81×/1.43× inference acceleration.

A common solution for deep learning application on devices is to leverage the high computational power by offloading the inference tasks to the server [6]. Yu [7] adopted a transfer learning method combined with ResNet [8] to achieve a recognition accuracy of 91.51%. Valeria [9] conducted a comparative fine-tuning study on mainstream CNN architectures, achieving an AUC of 99.72%.

Edge-cloud collaborative inference reduces latency and energy consumption by partitioning DNN models, unlike cloud-only approaches that suffer from high data transmission demands. Lin [10] proposed Edgent, a bandwidth-aware adaptive DNN partitioning framework for edge-cloud collaboration to reduce latency by leveraging edge computation. Liu [11] designed a DNN layer latency predictor for heterogeneous platforms, achieving a 20.81% average inference speedup. Gao [12] developed a triple-partition network with entropy-TOPSIS optimization, cutting end-to-end latency by over 3× without accuracy loss.

## 3 Methodology

### 3.1 Proposed Framework

The proposed framework utilizes partitioning to split the DNN model by using smaller intermediate layer outputs compared to raw input data. Front layers requiring less computation but more data transfer run on edge devices, while back layers needing heavy computation but less communication are processed on the cloud. This reduces latency, energy consumption, and bandwidth usage, balancing computational and communication demands.

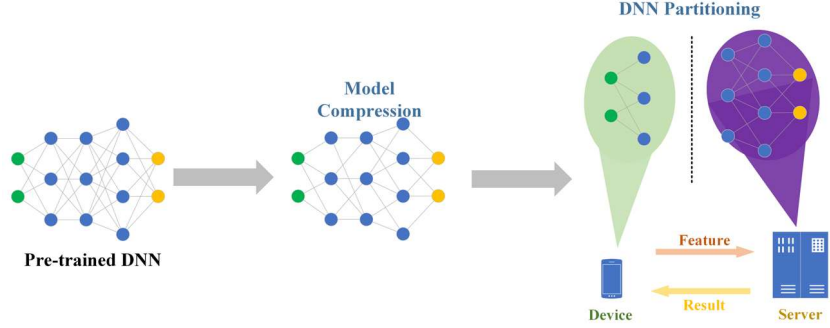

**Fig. 1.** Proposed Wireless Collaborated Inference Acceleration Framework

The framework accelerates collaborative inference through two core stages: (1) DRL-based layer-wise sparsity optimization for model compression and (2) latency-aware greedy search for cloud-edge partition point selection. As depicted in Fig.1, a pre-trained model is first compressed via reinforcement learning to determine layer-specific pruning strategies, then partitioned at the latency-optimal split point using greedy evaluation. The resulting edge-side and cloud-side submodels are deployed accordingly, achieving minimal end-to-end latency while balancing computational and communication resources.

### 3.2 Model Compression

Model compression employs deep reinforcement learning to make decisions about pruning strategies at the layer level, which automatically determines the optimal sparsity ratio for each layer while pruning the model without sacrificing accuracy.

The AMC method [5] will be utilized for automated pruning. Deep Deterministic Policy Gradient (DDPG) [13] is employed to determine the pruning ratio for each layer. The state space represents the space in which the problem is solved, and here the environment is the parameters of each network layer. For each network layer $i$, its state $s_i$ can be described as follows

$$(i, n, c, h, w, stride, k, FLOPs[i], F_{rdc}, F_{rest}, a_{i-1}) \qquad (1)$$

where $i$ denotes the layer index; $n$ and $c$ represent the number of output and input channels, respectively, $h$ and $w$ denote the height and width of the feature map, *stride* and $k$ denote the step size and convolutional kernel size. $FLOPs[i]$ denotes the floating-point computation in layer $i$, $F_{rdc}$ and $F_{rest}$ represent the floating-point computation reduced and remaining of layer $i$, and $a_{i-1}$ denotes the previous action taken by layer $i$.

The action space is the sparsity ratio of each convolutional layer with a continuous action space $a \in (0,1]$.

The reward function is defined as $r = Acc$, where $Acc$ represents the accuracy of the model.

During the strategy update training process, the transfer state is ($s_i$, $a_i$, $r_i$, $s_{i+1}$) in each round, where $r$ is the reward after the network is pruned. Based on the Bellman equation, the loss function in training is defined as Eq. 2.

$$Loss = \frac{1}{N}\sum_{i=1}(y_i - Q(s_i, a_i|\theta^Q))^2 \quad (2)$$

Where $y_i$ is defined as follows:

$$y_i = r_i - b + \gamma Q(s_{i+1}, \mu(s_{i+1})|\theta^Q) \quad (3)$$

Where the baseline reward $b$ is subtracted to reduce the variance of the gradient estimate and the discount factor γ is set to 1 to avoid over-prioritizing short-term rewards.

To better explore the action space, we use a truncated normal distribution to add some random noise in the strategy output, whose expression goes into Eq. 4.

$$\mu'(s_i) \sim TN(\mu(s_i|\theta_i^\mu), \sigma^2, 0.1) \quad (4)$$

where the noise σ is initialized to 0.5 and decays exponentially after each round.

### 3.3 DNN Partitioning

This section analyzes the data size and processing latency of each layer in AlexNet during the execution of forward inference.

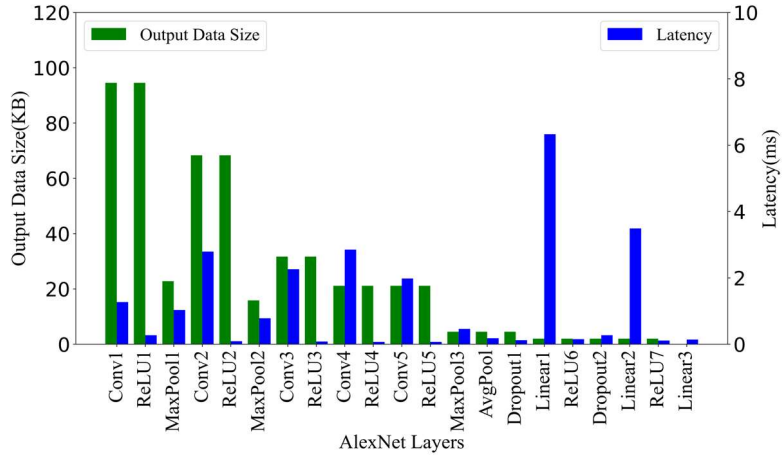

**Fig. 2.** Layer-wise Output Data Size and Delay

The model accepts a raw plant disease image sized 16.50KB with a resolution of 256×256 pixels. After data pre-processing, the image is converted to a 1×3×224×224 format, resulting in a file size of 73.50KB. The output data size and delay for each layer are shown in Fig. 2, where the green and blue bars represent the output data size and delay after processing by different layers, respectively. Output data size decreases as the number of layers increases. After passing through the first two convolutional layers, Conv1 and Conv2, the data size increases rapidly. After the maximum pooling layer, the data size decreases significantly, as the pooling layer effectively reduces the dimensionality of the data. Subsequently, after passing through the deeper fully connected layers, the output data size decreases continuously.

Based on the above analysis, an ideal split point can be identified among the candidate split points of AlexNet, contributing to minimizing the overall inference latency and reducing communication overhead.

Model deployment can be performed when the optimal split point is selected. The DNN model is divided into two parts, the device-side model and the server-side model, which are separated at the optimal split point. The two parts are deployed on the edge device and the cloud server respectively. The edge device conducts forward propagation by computing the device-side model layer by layer on the basis of the input data. The intermediate features are transmitted to the cloud server through the socket protocol, and the cloud server processes the server-side model layer by layer. Finally, the cloud server returns the inference results to the edge device.

### 3.4 Problem Formulation

This section will analyze the collaborative inference process in detail and then establish the latency model.

Collaborative inference latency includes device computation latency, server computation latency and intermediate transmission latency. In the collaborative inference process, the computing time and the corresponding latency for the layer-by-layer computation at the device and server, is device computation latency and server computation latency respectively. The transmission latency is the time required to transmit the intermediate features to the server side, so the collaborative inference latency can be calculated as Eq. 5.

$$T = T_D + T_{TX} + T_S \qquad (5)$$

Where $T_D$, $T_S$ and $T_{TX}$ represent the device computation delay, server computation delay, and intermediate feature variable transmission delay respectively.

The optimization problem is formalized as: given a neural network model $G$ with model parameters $\theta$, find the corresponding co-inference split point $c$ and the optimal pruning strategy $S$ for the minimum inference latency, which can be described as Eq. 6.

$$\underset{c,S}{argmin}\, T(G(\theta), c) = \underset{c,S}{argmin}(T_D + T_{TX} + T_S) \qquad (6)$$

where $S = \{S(l) | l \leq N_M, l \in N_+\}$, $1 \leq c \leq N, c \in N_+, 0 \leq S(l) \leq 1, l \leq N_M, l \in N_+$. $N$ is the maximum number of layers of the selected model and $S(l)$ represents the lth sparsity ratio for layers.

We design it as a joint optimization problem of the wireless collaborative reference split point $c$ and the optimal pruning strategy $S$. The selection of appropriate parameters for DNN partitioning and model compression is critical to achieving optimal inference acceleration. However, this constitutes a nonlinear mixed-integer programming problem involving multiple layer-wise sparsity ratios and split point, which spans an enormous solution space. To address this challenge, we propose a novel two-stage optimization approach capable of determining the optimal strategy.

### 3.5 Algorithm

The pseudo-code to solve the aforementioned optimization problem is illustrated in Algorithm below:

---
**Algorithm 1** Proposed Wireless Collaborated Inference Acceleration Algorithm
---
1: **Input**: Model $G(\theta)$
2: **Output**: Optimal pruning strategy $S$ and split point $c$
3: Randomly initialize Critic-Network $Q$ and Actor-Network $u$
4: Initialize target networks $Q'$ and $u'$
5: Create experience replay buffer $R$
6: **For** $ep = 1$ to $E_{max}$:
7:     Obtain initial observation state $s_1$
8:     **For** $t = 1$ to $N$:
9:         Select action $a_t$ using current policy $u$
10:        Execute action $a_t$
11:        Receive reward $r_t$ and next state $s_{t+1}$
12:        Store transition $(s_t, a_t, r_t, s_{t+1})$ in $R$
13:        Sample $N$ transitions $(s_i, a_i, r_i, s_{i+1})$ from $R$
14:        Compute $y_i$ using Eq. (3)
15:        Update Critic-Network by minimizing loss (Eq. 2)
16:        Update Actor-Network via policy gradient
17:        Update target networks $Q'$, $u'$
18: Obtain optimal pruning strategy $S$
19: Prune the model to get $G'(\theta')$
20: Set $T_{min} = T(G'(\theta'), 1)$, $c = 1$
21: **For** $j = 2$ to $N$:
22:     Compute $T(G'(\theta'), j)$ via timestamps
23:     $T_{tmp} = T(G'(\theta'), j)$
24:     If $T_{tmp} < T_{min}$:
25:         $c = j$
26:         $T_{min} = T_{tmp}$
27: Return optimal split point $c$

---

The DNN model for plant disease recognition undergoes pruning via deep reinforcement learning to enhance inference speed. Subsequently, an optimal split point is determined through a greedy algorithm to achieve maximally accelerated collaborative inference.

## 4 Experiment

### 4.1 Setup

**Dataset.** The Plant Village [14] dataset is a large-scale open-source image dataset constructed by Pennsylvania State University for plant disease research. It contains 54,305 images of diseased leaves, classified into 38 categories. Each image is uniformly sized at 256×256 pixels and is in JPG format. We apply a hierarchical partitioning method to perform intra-class stratification for each category to ensure an even distribution of samples in the training and test sets. The samples for the 38 categories of diseases in each category are divided into the training set at a ratio of 80%, and the remaining 20% of the samples are divided into the test set.

**Environment Configuration.** In this experiment, the edge device is configured with an Intel Core i7-6700 CPU (4 cores, 3.4GHz, 8G RAM). The server is equipped with an AMD Ryzen 5 5600 6-Core Processor (64G RAM, 2200MHz) and an NVIDIA GeForce RTX 3090 GPU (24GB VRAM). Both sides have an environment based on Python 3.8.18, Pytorch 2.0.0, and CUDA 12.2.

The optimization scheme employs Stochastic Gradient Descent (SGD) with momentum, where the initial learning rate is set to 0.01 and the momentum coefficient is configured at 0.9 to accelerate the model convergence process. A StepLR learning rate scheduler is implemented to execute a stepwise decay strategy, specifically designed to multiply the learning rate by a decay factor of 0.1 (gamma=0.1) every 20 training epochs (step_size=20). Batch size is set to 32.

### 4.2 Results and Analysis

**Model Compression.** The target sparsity ratio is set to 20%. Both the Actor and Critic networks are designed with two hidden layers, each containing 300 neurons. The replay buffer size is set to 500 transitions. For the first 100 pruning training iterations, a fixed noise level of $\sigma = 0.5$ is used for action exploration, which then decays exponentially in subsequent iterations. As shown in Fig. 3, the sparsity ratios of the five convolutional layers are 1, 0.875, 0.125, 0.292, and 0.313, respectively.

Fig. 4. indicates that the pruned model outperforms the original model in both aspects. Specifically, except for Conv1, the output data sizes are reduced by approximately 12.51%, 87.50%, 70.84%, and 68.75% in each layer after pruning. This demonstrates that the pruned model effectively reduces the number of parameters and storage requirements. Besides, the processing latencies of each layer are also reduced after pruning, with decreases of 88.17%, 93.39%, 91.93%, and 80.30%, respectively. This highlights that the pruned model accelerates the inference speed, enhancing the user experience.

Since the accuracy of the model often decreases after pruning, fine-tuning is essential to recover the lost accuracy. It can be seen in Table 1 that the accuracy of Top k of the pruned model only decreases by 0.91%, 0.15%, and 0.07%, respectively, compared to the original model; and the Top-k accuracy of the fine-tuned model increases by 4.41%,

0.6% and 0.26%, respectively, compared to the pruned model. The proposed model compression method effectively enhances the inference speed under the premise of good performance.

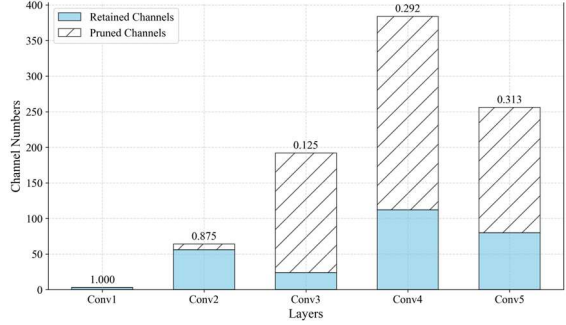

**Fig.3.** Channel Numbers of Layers

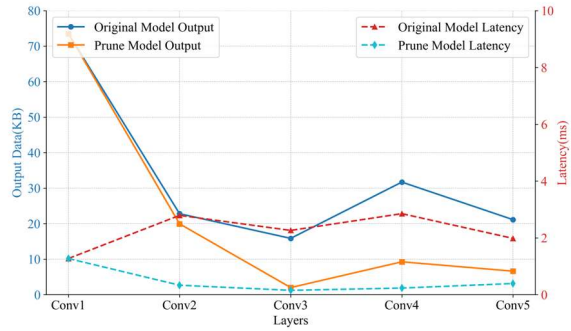

**Fig.4.** Layer-Wise Output Data Size and Latency

Table 1. Top-k accuracy

| Model Type | Top-1 accuracy | Top-3 accuracy | Top-5 accuracy |
|---|---|---|---|
| Original model | 93.67% | 99.32% | 99.77% |
| Pruned model | 92.76% | 99.17% | 99.70% |
| Fine-tuned model | 97.17% | 99.77% | 99.96% |

**DNN Partitioning.** The impact of split point selection in DNN partitioning is evaluated under a Wi-Fi network environment ten times with a bandwidth of nearly 50 Mbps. As shown in Table 2, the collaborative inference latency reaches its minimum at split point 6, indicating that it is the optimal one in this study. It is close to the input data, allowing inference to begin earlier without waiting for extensive intermediate results. Moreover, a MaxPool layer reduces the size of the output data before this point, significantly decreasing communication overhead.

Table 2. Split points and corresponding latency

| Split Point | 1 | 2 | 3 | 4 | 5 |
|---|---|---|---|---|---|
| Latency | 99.91 | 166.98 | 65.89 | 85.03 | 31.91 |
| **Split Point** | 6 | 7 | 8 | 9 | 10 |
| Latency | 20.07 | 60.88 | 40.98 | 55.93 | 37.96 |
| **Split Point** | 11 | 12 | 13 | 14 | 15 |
| Latency | 57.79 | 36.11 | 27.96 | 26.34 | 39.15 |
| **Split Point** | 16 | 17 | 18 | 19 | 20 |
| Latency | 34.57 | 31.75 | 36.04 | 36.67 | 36.59 |

**Co-inference Acceleration.** To demonstrate efficiency, the proposed framework is referred to as the pruned_co_infer method, and compared with baseline methods in terms of latency. As shown in Fig.5, the device-only method deploys the entire model on the edge device with an average latency of 31.36ms. The server-only method requires all input data to be transmitted. Therefore, the inference latency depends on the wireless transmission bandwidth, with an average latency of 80.78ms. Both the device-only and the server-only methods can achieve acceleration through pruning compared to their original forms. The pruned_co_infer method achieves an average latency of 18.55ms. Compared to the edge-only and cloud-only methods, it achieves a 1.69× and 4.35× speedup, respectively, demonstrating the efficiency of collaborative inference.

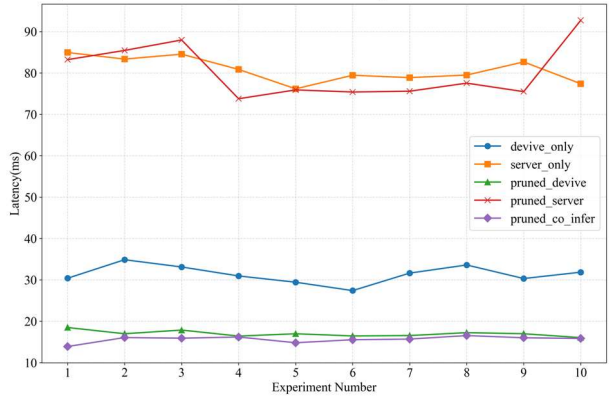

Fig. 5. Comparison of Different Approaches

### 4.3 Implementation

To enhance user-friendly interaction with computers, we developed a socket-based wireless collaborative inference application system to recognize plant diseases using Gradio [15], a well-known web UI framework released by Hugging Face.

The trained model is deployed on both the edge devices and the cloud server. On the edge client, the system invokes the initial layers of the DNN model to process disease images in real-time, generating intermediate feature results and sending them to the cloud server via socket protocol. On the cloud server side, it processes the intermediate results using the later layers of the DNN model and returns the inference results to the edge client for further use.

The main functional modules in the system include parameter setting, recognition results, model architecture demonstration with split point, and prevention suggestion. An example of the system's Graphical User Interface (GUI) is shown in Fig.6.

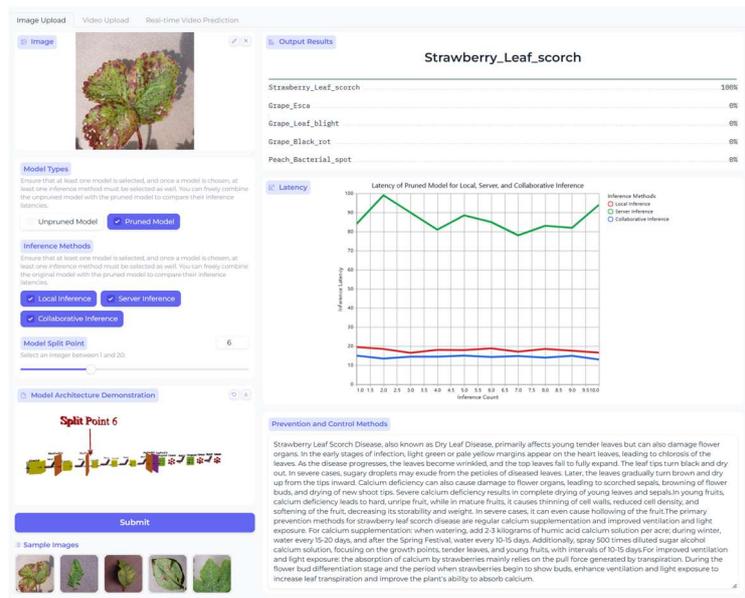

**Fig. 6.** System GUI of Wireless Collaborated Inference for Plant Disease Recognition

As demonstrated in Fig.6, the system provides recognition results for plant diseases from user-provided images, videos, or real-time video streams. It utilizes the Model3D module in Gradio to enable interaction with the deep learning model stored in GLB format, and it displays both the model structure and the split point. The latency comparison curves of real-time collaborative inference and the baseline method are illustrated under user parameter settings for data visualization. Additionally, the database matching queries provide suggestions for corresponding disease prevention.

## 5  Conclusion

In conclusion, this paper presents a novel wireless collaborative inference acceleration framework, which addresses the critical challenges of deploying deep learning-based applications on hardware devices with limited resources for plant disease recognition.

Our approach leverages a DDPG-based method to automates layer-wise sparsity allocation and latency-minimized cloud-edge partitioning through a two-stage process: 1) reinforcement learning-guided pruning policy generation and 2) greedy algorithm-based model splitting. Experimental results show that the framework greatly improves inference speed while keeping recognition accuracy at a satisfactory level. Furthermore, the implementation includes a Gradio-powered interface for real-time plant disease diagnosis, demonstrating practical viability in smart agriculture applications.

**Acknowledgement.** #The two authors contributed equally to this work. This research was supported by the Sichuan Science and Technology Program under Grant 2023YFG0302 and the Fundamental Research Funds for the Central Universities, Southwest Minzu University under grant ZYN2023098.